\documentclass[10pt,twocolumn,twoside]{IEEEtran}
\ifCLASSOPTIONcompsoc
  \usepackage[compress]{cite}
\else
  \usepackage[compress]{cite}
\fi
\usepackage{amsmath}
\usepackage{amsfonts}
\usepackage{amssymb}
\usepackage{math}
\usepackage{graphicx}

\usepackage{url}
\usepackage{balance}
\usepackage{multirow}
\usepackage{tablefootnote}
\usepackage{algorithm}
\usepackage{algpseudocode}
\usepackage{xcolor}
\usepackage{tikz}
\usetikzlibrary{bayesnet,calc}

\newcommand{\fref}[1]{Fig.~\ref{#1}}

\newcommand{\fv}[1]{#1^{\textsc{f}}}
\newcommand{\bv}[1]{#1^{\textsc{b}}}
\newcommand{\iv}[1]{#1^{\textsc{i}}}
\newcommand{\jv}[1]{#1^{\textsc{j}}}

\newcommand{\fg}[1]{\tilde{#1}^{\textsc{f}}}

\usepackage[final]{review}
\setcoverletter{coverletter-R1.tex}
\setrevision{1}

\title{Variational Inference and Learning of\\ Piecewise-linear Dynamical Systems}
\author{
        Xavier~Alameda-Pineda,~\IEEEmembership{Senior Member,~IEEE},
        Vincent~Drouard~
         and~Radu~Horaud
\thanks{X. Alameda-Pineda and R. Horaud are with INRIA Grenoble Rh\^one-Alpes, Montbonnot Saint-Martin, France.}
\thanks{V. Drouard is with Image Metrics, Manchester, UK.}
\thanks{This work is supported by the Multidisciplinary Institute in Artificial Intelligence (MIAI), Grenoble, and funded by the ANR under grand agreement ANR-19-P3IA-0003; as well as by the ML3RI project of the Young Researcher (Jeunes Chercheuses et Jeunes Chercheurs) programme of the ANR, under grant agreement ANR-19-CE33-0008-01.}}

\begin{document}
\maketitle
\begin{abstract}
Modeling the temporal behavior of data is of primordial importance in many scientific and engineering fields. Baseline methods assume that both the dynamic and observation equations follow linear-Gaussian models. However, there are many real-world processes that cannot be characterized with a single linear behavior.  Alternatively, it is possible to consider a piecewise-linear model which, combined with a switching mechanism, is well suited when several modes of behavior are needed. Nevertheless, switching dynamical systems are intractable because of their computational complexity increases exponentially with time.
In this paper, we propose a variational approximation of piecewise linear dynamical systems. We provide full details of the derivation of two variational expectation-maximization algorithms, a filter and a smoother. 
We show that the model parameters can be split into two sets, static and dynamic parameters, and that the former parameters can be estimated off-line together with the number of linear modes, or the number of states of the switching variable. We apply the proposed method to a visual tracking problem, namely head-pose tracking, and we thoroughly compare our algorithm with several state of the art trackers. 
\end{abstract}


\section{Introduction}
\label{sec:introduction}
\addnote[temp-ex]{1}{Modeling the temporal behavior of data is of primordial importance in many scientific fields, such as signal processing \cite{wu2004modeling}, computer vision \cite{ba2016line,byeon2019variational}, robotics \cite{thrun2006probalistic}, autonomous navigation \cite{kooij2019context}, to cite just a few}. The baseline model for addressing this problem  is the linear dynamical system (LDS). The basic idea of LDS is to assume that both the dynamic and the observation equations follow linear-Gaussian models. This yields tractable learning and inference procedures, namely the Kalman filter (KF) \cite{kalman1960new} and the Kalman smoother (KS) \cite{rauch1965maximum}. 

In many situations, the latent (state) space, whose dynamics must be modeled, is embedded in a high-dimensional observed space. In general, the direct mapping, from the  low-dimensional latent space to the high-dimensional feature (observed) space, as well as the inverse of this mapping, are non-linear. Moreover, the dynamics of the latent space may be non-linear as well. Several methods were proposed to deal with non-linear dynamical systems, e.g. Bayesian tracking with particle filters \cite{chen2000mixture,arulampalam2002tutorial}, the extended Kalman filter (EKF) \cite{jazwinski2007stochastic}, and the unscented Kalman filter (UKF) \cite{julier2004unscented}. Alternatively, it is possible to consider several linear dynamic models and to combine them with a switching mechanism that selects over time one among several linear regimes: this is referred to as the switching Kalman filter (SKF), the switching LDS, the jump-Markov linear system, or the switching state space model \cite{murphy2012machine}. 


The mixture Kalman filter (MKF)  \cite{chen2000mixture} uses a sequential Monte Carlo method based on a random mixture of Gaussians to approximate the target distribution. It formulates non-linear systems into conditional or partial conditional LDSs. Outcomes of non-linear/non-Gaussian Bayesian trackers based on sequential importance sampling are reviewed and discussed in \cite{arulampalam2002tutorial}, most notably the problems of degeneracy, choice of importance density, and resampling.  The basic idea of EKF is to linearize the equations using a first-order Taylor expansion and to apply the standard KF to the linearized model. The additional error due to linearization is not taken into account which  may lead to sub-optimal performance. Rather than approximating a non-linear dynamical system with a linear one, UKF specifies the state distribution using a minimal set of deterministically selected sample points. The sample points, when propagated through the true non-linear system, capture the posterior state distribution accurately to the third order Taylor expansion. 
The stability of UKF was thoroughly investigated in the control literature, e.g.  \cite{xiong2006performance}. It was shown that the design of the noise covariance, of both state and observation equations, critically affects the performance of the filter. 

The methods just outlined generally deal with a single non-linear or linearized, state equation. There are many real-world processes that cannot be characterized  with a single state equation, but with multiple discrete modes of behavior, both in terms of their dynamics and of their observation model, in particular when the latter must predict (generate) high-dimensional observations from a low-dimensional state space. 

We consider the problem of tracking the orientation of a person head/face (three rotation angles) from a sequence of images, referred to as \textit{head-pose tracking} (HPT). A face detector provides input to a face descriptor immune to illumination changes, background conditions, as well as inter- and intra-person variabilities (shape and aspect). Face descriptors of choice are histograms of oriented gradients (HOGs)  \cite{dalal2005histograms} and embeddings based on convolutional neural networks (CNNs) \cite{ahn2014real,mukherjee2015deep,ranjan2017hyperface,lathuiliere2017deep}. These high-dimensional feature vectors contain head pose information implicitly and a number of non-linear or piecewise-linear regression methods have been proposed to extract head pose, namely Gaussian process regression \cite{Rasmussen06gaussianprocesses}, support vector regression \cite{smola2004tutorial}, kernel partial least squares \cite{abdi2003partial},  deep inverse regression \cite{lathuiliere2017deep}, or Gaussian mixture of linear regressions \cite{DeleforgeForbesHoraud2015}, \cite{drouard2017robust}.

Recently it was proposed to approximate non-linear high-dimensional to low-dimensional (high-to-low) mappings with mixtures of linear-Gaussian \cite{DeleforgeForbesHoraud2015}, \cite{tu2019prediction} and linear-Student  \cite{perthame2016inverse} regressions. These models adopt an \textit{inverse regression} strategy, namely they learn a low-to-high mapping followed by the evaluation of a high-to-low mapping. The rationale of this way of doing is manyfold:  (i)~low-to-high regression learning avoids the estimation of a large number of parameters, hence it requires a relatively small amount of training data, (ii)~the parameters of the high-to-low regression are analytically evaluated from the low-to-high parameters, (iii)~the mixture model setting has the advantage of providing inference procedures using closed-form EM algorithms. It is interesting to note that these Gaussian/Student mixtures group data with similar regression associations together. Within the same
cluster, the association can be considered as locally linear, which can then be resolved
under the classical linear regression setting. This \textit{piecewise linear} models are well suited to capture potentially non-linear relations. 
This was extensively discussed in \cite{DeleforgeForbesHoraud2015} and in \cite{tu2019prediction}, and was successfully applied to both head-pose estimation \cite{drouard2017robust} and audio-source localization \cite{deleforge2015co,li2016estimation}.

\addnote[plds]{1}{In this paper we propose a variational expectation-maximization algorithm to learn \textit{piecewise-linear dynamic systems} (P-LDSs). A P-LDS can be viewed either as a piecewise-linear approximation of a non-linear dynamic system \cite{arulampalam2002tutorial}, or as a dynamic generalization of mixtures of linear regressions \cite{DeleforgeForbesHoraud2015,deleforge2015co,drouard2017robust}.}
A P-LDS may also be viewed as soft version of switching LDS \cite{ghahramani2000variational,murphy2012machine}.
The assignment variable of the piecewise-linear mixture model plays the role of the switching variable of both the dynamic and observation models and it is governed by the transition matrix of a corresponding hidden Markov model (HMM). It is well known that the complexity of these hybrid dynamical systems, i.e. systems that combine discrete- and continuous-valued latent variables, increases exponentially with time \cite{murphy2012machine}. Indeed, for $K$ linear models and after $T$ time steps, the exact marginalized posterior distribution of the state is a Gaussian mixture with $K^T$ components. Therefore, the problem of learning the parameters of such hybrid systems must be carried out via approximate solutions. Traditionally, inference of hybrid dynamical systems is based on approximating the posterior distribution with a simpler one, e.g. the generalized pseudo Bayes filter. In this paper we propose an alternative based on replacing the difficult-to-compute posterior with an approximate tractable posterior. 

The remainder of this paper is organized as follows. Section~\ref{sec:related} describes related work. Section~\ref{sec:temporal} formulates P-LDS and analyses its intractability. Section~\ref{sec:estimation} describes in detail the proposed variational approximation model as well as the as two EM algorithms, a variational filter and a variational smoother. For the sake of completeness, Section~\ref{section:gpb2} describes a GPB2  approximation of P-LDS. Section~\ref{sec:experiments} describes experimental results obtained with head-pose tracking. 
\footnote{Supplemental materials can be found at \url{https://team.inria.fr/perception/research/learning-plds/}.}

\section{Related work}
\label{sec:related}

The intractability of switching LDS (and of P-LDS) can be addressed using sampling methods: sequential Monte Carlo methods (particle filtering) have been used for this purpose. The main drawback is that sampling can be inefficient, which leads to slow convergence. To reduce the size of the state space, Rao-Blackwellisation may be employed. Instead of drawing the samples from the joint posterior of the discrete and continuous states, tractable sub-structures of the model can be utilized. \addnote[nonpar-bayes]{1}{Non parametric Bayesian inference of switching LDS was proposed in \cite{fox2011bayesian}, where a switching mechanism is used for the dynamic model, while the observation model uses a standard LDS. This is problematic when the observation model is not linear. Moreover, a Gibbs sampler is used for inference, providing asimptotic properties tied to a high computational cost.}

A general theory
of Rao-Blackwellised particle filters (RBPFs) applied to dynamic Bayesian networks (DBNs) was proposed in \cite{doucet2000rao}. In the case of switching LDS, marginalization of the joint posterior, namely analytic integration over the continuous variables, considerably reduces the sampling space. RBPF using Gibbs sampling was used for speech recognition \cite{rosti2004rao}, where the discrete state corresponds to phonemes and the continuous state corresponds to a time evolving representation of the observations. RBPF using Metropolis-Hastings sampling was used in \cite{oh2008learning} to analyse motion patterns of bees. 

In  \cite{oh2008learning} it was noticed that a naive exploration of the space of discrete variables is prohibitive and that data-driven (DD) MCMC sampling improves convergence. DD-MCMC requires supervised learning from a labeled training dataset, which may be cumbersome, if not prohibitive, because it is not practical to manually associate discrete-variable values with the observed vectors. More recently,  \cite{kooij2016mixture} proposed a mixture of switching LDSs to analyse the dynamic behavior of pedestrians: an MCMC inference scheme uses both Gibbs and Metropolis-Hastings samplers. While potentially powerful, MCMC methods and their variants are non-analytic methods and typically suffer from slow convergence rates (they are only exact in the case of infinite size samples), especially in high-dimensional spaces, which is impractical in the case of tracking. A recent comparison between MCMC and variational inference emphasizes that the latter easily takes advantage of stochastic and distributed optimization \cite{blei2017variational}.

The \textit{generalized pseudo Bayes} (GPB) \cite{bar1993estimation} and the GPB of order two (GPB2) \cite{bar1988tracking}, \cite{murphy2002dynamic} algorithms belong to the \textit{assumed density filtering} (ADF) \cite{Boyen1998tractable} class of models which is widely used to
approximate an intractable distribution with a simpler one. GPB collapses the mixture of $K$ Gaussian components, resulting from considering all the possible linear models at each time step, into one Gaussian component. GPB2 is more sophisticated and more time consuming than GPB, as it 
collapses the $K^2$ Gaussian components, resulting from considering all the possible linear models when going from one time step to the next, into $K$ Gaussians components. The GPB2 algorithm was applied to the analysis of motor cortical activity of hand movements in macaque monkeys \cite{wu2004modeling}, and more recently it was used for tracking eye gaze \cite{masse2018tracking} and for path prediction of pedestrians in the context of intelligent vehicles \cite{kooij2019context}. An offline extension of GPB2, called expectation correction was proposed in \cite{barber2006expectation} and applied to speech recognition robust to noise \cite{mesot2007switching}.

Alternatively, structural variational inference and learning techniques consider a parameterized distribution which is in some sense close to the desired posterior distribution, as well as easier to compute. Variational models modify the structure of the posterior by removing dependencies between variables, i.e. the joint posterior distribution $P$ is approximated by a tractable \textit{variational} distribution $Q$ with variational parameters $\thetavect$. The Kullback-Leibler divergence between $Q$ and $P$ is minimized with respect to $\thetavect$. In the case of switching LDSs, \cite{pavlovic1999variational}, \cite{ghahramani2000variational} and \cite{lee2003variational} propose to remove some of the dependencies between the continuous and discrete latent variables, thus yielding tractable solvers. The mixed-state DBN proposed in \cite{pavlovic1999variational} is an HMM driving the LDS bias. In \cite{ghahramani2000variational} an HMM switches between several LDSs, each LDS having a different latent variable with its own dynamic regime. Both \cite{pavlovic1999variational} and \cite{ghahramani2000variational} lead to an EM algorithm whose maximization step (learning) satisfies a set of fixed-point equations in the variational parameters.  The variational model proposed in \cite{lee2003variational} is more general than the one proposed in \cite{pavlovic1999variational} and in \cite{ghahramani2000variational}. Nevertheless, their approximation breaks the dependencies between the HMM and the LDS as well as the temporal dependencies.

The variational model of  \cite{lee2003variational}  was applied to speech recognition \cite{lee2004multimodal} and to speech production \cite{deng2004switching}, while the 
model of \cite{pavlovic1999variational} was applied to human motion capture \cite{pavlovic2000learning}. 
It is interesting to note that in spite of the recent popularity gained by variational models, e.g. \cite{zhang2019advances}, as they provide tractable solutions to various intractable inference problems, e.g. \cite{ma2014variational,ma2014bayesian,taghia2014bayesian,taghia2015variational,deleforge2015hyper,ba2016line,ma2019insights,byeon2019variational,ban2019tracking,ban2019variational}, at the best of our knowledge, variational inference of switching LDS has not been addressed for the last decade.

\addnote[dgm]{1}{More recently, learning and inference of dynamical systems have been addressed in the framework of deep generative models (DGMs), where the linear-Gaussian transition and emission distributions of LDS are replaced with non-linear Gaussian models. In detail, the mean and covariance of a Gaussian distribution are modeled with neural networks. Because of this non-linear dependencies, direct optimization of the corresponding data log-likelihood function is intractable. This issue is solved by maximization of a variational lower bound of the  
log-likelihood. For example, \cite{krishnan2017structured} uses a recurrent neural network (RNN) to model the mean and diagonal covariance matrix. The proposed structured inference network corresponds to a deep Kalman smoother, that needs both past and future observations. This formulation belongs to a wider class of non-linear Gaussian state-space models that were recently reviewed in \cite{girin2020dynamical}.} 

With respect to the related work just outlined,
this paper has the following contributions. We propose a variational approximation of P-LDS. Unlike the variational model of \cite{ghahramani2000variational}, which switches between several linear regimes, we propose a piecewise-linear model that approximates non-linear dynamic models. Unlike \cite{lee2003variational}, the proposed variational approximation doesn't break the temporal dependencies. \addnote[nlgm]{1}{Unlike the RNN-based non-linear state-space model of \cite{krishnan2017structured}, the proposed variational piecewise-linear model yields a closed-form solver and it has fewer parameters}.

We provide full details of an EM algorithm that can be indifferently used either as a \textit{variational filter} or as a \textit{variational smoother}. Unlike the models of \cite{pavlovic1999variational} and of \cite{ghahramani2000variational} that lead to solving a set of fixed-point equations, we develop a closed-form solution for learning the model parameters. Moreover, the proposed method benefits from a closed-form EM algorithm for the off-line estimation of the static parameters, namely, those associated with the observation model. This has two practical outcomes: (i)~it reduces the task of learning to the estimation of the parameters of the dynamic model and (ii) it allows to learn the number of states of the switching variable (or, equivalently, the number of linear models) based on a model selection principle, i.e. the Bayes information criterion, \cite{DeleforgeForbesHoraud2015}, \cite{perthame2016inverse}. For the sake of completeness, we describe a GPB2 algorithm for P-LDS that slightly differs from the standard GPB2 for switching LDS in that it only needs to estimate the parameters of the dynamic model.

\section{Problem Formulation}
\label{sec:temporal}
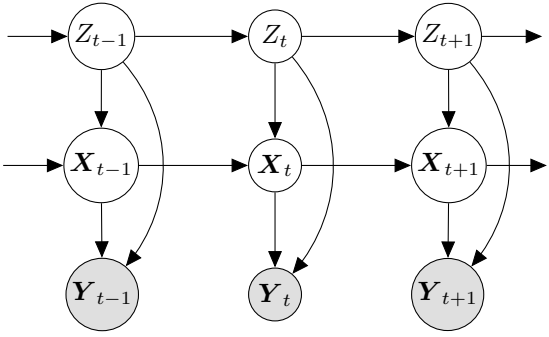
\begin{figure}
	\centering
	\begin{tikzpicture}[->]
        \node[latent] (z) {$Z_{t}$};
        \node[latent,left=of z,xshift=-5mm] (zm) {$Z_{t-1}$};
        \node[left=of zm,xshift=2mm] (zmm) {};
        \node[latent,right=of z,xshift=5mm] (zp) {$Z_{t+1}$};
        \node[right=of zp,xshift=-2mm] (zpp) {};
        \node[latent,below=of z] (x) {$\Xvect_{t}$};
        \node[latent,left=of x,xshift=-4.5mm] (xm) {$\Xvect_{t-1}$};
        \node[left=of xm,xshift=2mm] (xmm) {};
        \node[latent,right=of x,xshift=4.5mm] (xp) {$\Xvect_{t+1}$};
        \node[right=of xp,xshift=-2mm] (xpp) {};
        \node[obs,below=of x] (y) {$\Yvect_{t}$};
        \node[obs,left=of y,xshift=-4.5mm] (ym) {$\Yvect_{t-1}$};
        \node[obs,right=of y,xshift=4.5mm] (yp) {$\Yvect_{t+1}$};
        \edge{zmm} {zm};
        \edge{zm} {z};
        \edge{z} {zp};
        \edge{zp} {zpp};
        \edge{xmm} {xm};
        \edge{xm} {x};
        \edge{x} {xp};
        \edge{xp} {xpp};
        \edge{zm} {xm};
        \edge{z} {x};
        \edge{zp} {xp};
        \edge{xm} {ym};
        \edge{x} {y};
        \edge{xp} {yp};
        \path(zm) edge[bend left=40] (ym);
        \path(z) edge[bend left=40] (y);
        \path(zp) edge[bend left=40] (yp);
	\end{tikzpicture}
	\caption{Graphical representation of the switching linear dynamic model that show the dependencies between the latent variables (white circles) and the observed variables (gray circles).}
	\label{fig:model}
\end{figure}

Let $\Xvect\in\mathcal{X}\subset\mathbb{R}^L$ be a latent (or state) random variable and $\Yvect\in\mathcal{Y}\subset\mathbb{R}^D$ be an observation variable. Without loss of generality it will be assumed that the dimensionality of the observation space is much higher than the dimensionality of the latent space, $D\gg L$. Let $\xvect$ and $\yvect$ denote realizations of $\Xvect$ and $\Yvect$, respectively.
Let $t\in\mathbb{N}$ be the discrete time index: $\Xvect_t$ denotes the latent variable at $t$ and the notation $\Xvect_{1:t}$ is a shorthand for the temporal sequence $\Xvect_1,\dots,\Xvect_t$. In an LDS, the observed vectors are connected to the latent vectors through an observation equation. We will consider the following piecewise-linear observation model. It is assumed that at any time $t$ a realization $(\yvect_t,\xvect_t)$ of $(\Yvect_t,\Xvect_t)$ is such that $\yvect_t$ is generated from $\xvect_t$ by a linear function $\yvect = l_k(\xvect)$ plus an error term. At each time step $t$, a discrete latent variable $Z_t$ is introduced, such that $Z_t = k$ if and only if $\yvect_t$ is the image of $\xvect_t$ by $l_k$, with $k\in\{1,\dots, K\}\subset\mathbb{N}$. 
The \textit{piecewise} linear function that maps the state $\Xvect_t$ onto the observation $\Yvect_t$ is:
\begin{align}
\label{eq:obs}	
\yvect_t = \sum_{k=1}^{K} \mathbb{I}(Z_t=k) (\Amat_{Z_t} \xvect_t + \bvect_{Z_t} + \evect_{Z_t}),
\end{align}
where matrix $\Amat_{Z_t=k}\in\mathbb{R}^{D\times L}$ and vector $\bvect_{Z_t=k}\in\mathbb{R}^{D}$ define $l_k$, $\mathbb{I}(\cdot)$ is the indicator function and $\evect \sim \mathcal{N}(0,\Sigmamat)$ is a zero-mean Gaussian noise vector with covariance matrix  $\Sigmamat \in\mathbb{R}^{D\times D}$. The description of the model is completed by a similar piecewise linear dynamic equation:
\begin{align}
\label{eq:dyn}
\xvect_t = \sum_{k=1}^{K} \mathbb{I}(Z_t=k) (\Cmat_{Z_t} \xvect_{t-1} + \wvect_{Z_t}),
\end{align}
where $\Cmat_{Z_t}\in\mathbb{R}^{L\times L}$ is the state transition matrix and  $\wvect \sim \mathcal{N}(0,\Qmat)$ is a zero-mean Gaussian noise vector with covariance matrix  $\Qmat \in\mathbb{R}^{L\times L}$. To summarize, the $k$-th LDS is defined by the following probabilistic model, see~\fref{fig:model}:
 \begin{align}
 \label{eq:dynamic}
 p(\xvect_t  | \xvect_{t-1}, Z_t=k) &= \mathcal{N} (\xvect_t; \Cmat_{Z_t} \xvect_{t-1}, \Qmat_{Z_t}),\\
 \label{eq:observation}
p(\yvect_t  | \xvect_t, Z_t=k) &= \mathcal{N} (\yvect_t; \Amat_{Z_t} \xvect_t + \bvect_{Z_t}, \Sigmamat_{Z_t}),\\
  \label{eq:model_pXZ}
 p(\xvect_1 |Z_1=k) &= \mathcal{N}(\xvect_1 ; \gammavect_{Z_1},\Gammamat_{Z_1}), \\
  \label{eq:prior_Z}
 p(Z_1=k) &= \pi_{Z_1},
 \end{align}
where $\{\gammavect_k, \Gammamat_k, \pi_k\}_{k=1}^K$, $\gammavect_k\in\mathbb{R}^{L}$, $\Gammamat_k\in\mathbb{R}^{L\times L}$ and $\pi_k$, define the Gaussian mixture parameters of the initial state. 

So far we did not specify the temporal behavior of the discrete hidden variables $Z_{1:t}$ that allow to select both the observation model \eqref{eq:obs} and the dynamic model \eqref{eq:dyn}. 
We assume that $Z_{1:t}$ obey a first-order Markov chain:
 \begin{align}
\label{eq:switching}
p (Z_t = i) &= \sum_{j=1}^{K}   p(Z_t=i | Z_{t-1}=j) p (Z_{t-1}=j) \nonumber \\
  \tau_{ij} &= p(Z_t=i | Z_{t-1}=j), \quad 1 \leq i,j \leq K
\end{align}
where $\tau_{ij}$ is an entry of a state transition matrix $\Tmat\in\mathbb{R}^{K \times K}$ which defines the switching behavior: at any time, the system evaluates a convex combination of $K$ Gaussian-linear observation models and dynamic regimes. 
The model above is described by the following parameters that we group into two sets: 
\begin{align}
\label{eq:theta}
\thetavect &= \lbrace\Amat_k, \bvect_k,\Sigmamat_k,\pi_k,\gammavect_k,\Gammamat_k\rbrace_{k=1}^{k=K},\\
\label{eq:phi}
\phivect &= \lbrace \Cmat_j, \Qmat_j, \tau_{ij} \rbrace_{i,j=1}^{i,j=K}.
\end{align}
\addnote[gllim-em]{1}{
The \textit{static} parameters $\thetavect$ in \eqref{eq:theta} 
characterize the observation model  \eqref{eq:obs} and \eqref{eq:observation}, the initial distribution of $\xvect$, \eqref{eq:model_pXZ}, and the prior \eqref{eq:prior_Z}: they do not depend on the dynamics of the sequence. Hence, $\thetavect$ can be learned from a training set of input-output pairs $\lbrace \yvect_n,\xvect_n\rbrace_{n=1}^N$ in the following way. Let $\mathbb{E}_{r_Z}[\log p(\yvect_{1:N}, \xvect_{1:N}, Z_{1:N}; \thetavect)]$ be the expected complete-data log-likelihood, where the expectation is taken over the responsibilities $r_Z=p(Z_{1:N} | \yvect_{1:N}, \xvect_{1:N} \thetavect)$. The parameter set $\thetavect$  can be estimated via a closed-form EM procedure, i.e. \cite{DeleforgeForbesHoraud2015}, that alternates between the evaluation of the responsibilities (posteriors) and the maximization of the expectation that was just defined. 
The number of linear models $K$, can be estimated via model selection, using the Bayes information criterion (BIC), or empirically, based on the mean absolute error between the predicted outputs and the ground-truth values, i.e \cite{drouard2017robust}. }

The parameters $\phivect$ characterize the \textit{dynamic} behavior of both the continuous \eqref{eq:dynamic} and discrete \eqref{eq:switching} state variables and hence they must be estimated from a temporal sequence of observations. It is interesting to note that when the observation space is of high dimension, the strategy consisting of independently estimating the parameters $\thetavect$ and $\phivect$ simplifies the tasks of P-LDS inference and learning by drastically reducing the number of parameters. Let, for example, $D=1000$ (the dimension of the observation space), $L=10$ (the dimension of the latent space), and $K=10$ (the number of linear-Gaussian models). If we assume diagonal covariance matrices $\Sigmamat_k$, $\mathrm{dim}(\thetavect)\approx 10^5$ and 
$\mathrm{dim}(\phivect) \approx 10^3$. 

\subsection{Computational Intractability}
\label{section:intractability}

Exact model inference, i.e. estimation of the dynamic parameters $\phivect$, is faced with an intractability problem. Indeed, let's analyze the complexity of computing the posterior distribution, namely the conditional probability of the state variable $\xvect_t$ given the present and past observations $p(\xvect_t | \yvect_{1:t})$. This distribution can be obtained by marginalization over the continuous and discrete variables given the observations:
\begin{align}
\label{eq:predictive-marginal}
p(\xvect_t | \yvect_{1:t} )= 
\sum_{z_{1:t}=1}^K 
\int_{\xvect_{1: t-1}} p(\xvect_t, \xvect_{1:t-1}, z_{1:t} | \yvect_{1:t}) d \xvect_{1: t-1},
\end{align}
where $\sum_{z_{1:t}=1}^K$, $\int_{\xvect_{1: t-1}}$ and $ d \xvect_{1: t-1}$ are shorthands for $\sum_{z_1=1}^K \dots \sum_{z_t=1}^K$, for $\int_{\xvect_1} \dots \int_{\xvect_{t-1}}$ and for $d \xvect_{1} \dots d \xvect_{t-1}$, respectively. Using the first-order Markov dependencies shown in the graphical model of \fref{fig:model}, the joint probability (right hand side of \eqref{eq:predictive-marginal}) can be factorized as:
\begin{align}
	p  ( & \xvect_{1:t}, z_{1:t} \vert \yvect_{1:t} ) \propto   p(\xvect_{1}, z_{1} \vert \yvect_{1} )  \nonumber \\
	\label{eq:joint_posterior_all} 
	 \times & \prod_{t'=2}^t  p\left( \yvect_{t'} \vert \xvect_{t'}, z_{t'} \right) p\left( \xvect_{t'} \vert \xvect_{t'-1}, z_{t'} \right) 
		  p\left(z_{t'} \vert z_{t'-1} \right).
\end{align}
Substituting the factors of \eqref{eq:joint_posterior_all} with their expressions \eqref{eq:dynamic}-\eqref{eq:switching} and using standard properties of Gaussian distributions and using marginalization, the joint distribution \eqref{eq:joint_posterior_all} writes:
\begin{align}
\label{eq:joint_posterior_all_final}
	p(\xvect_t, \xvect_{1:t-1}, z_{1:t} \vert \yvect_{1:t} ) = \beta_{t} \;  \mathcal{N}\left( \left[ \xvect_1 : \xvect_t\right]; \kappavect_{t} , \Kmat_{t}  \right),
	\end{align}
where  the notation $[\xvect_1 : \xvect_t]$ denotes  vertical concatenation of vectors $\xvect_{1}, \dots, \xvect_{t}$, and where the weight $\beta_{t} $,  mean $\kappavect_{t} $, and covariance $\Kmat_{t} $ depend on the model parameters \eqref{eq:theta} and \eqref{eq:phi} and on $Z_{1:t}$. Therefore, the predictive distribution 
\eqref{eq:predictive-marginal} is a Gaussian mixture with a number of components that increases exponentially with time, i.e. there are $K^t$ components after $t$ time steps, which is intractable.

This phenomenon appears not only when attempting to evaluate $p(\xvect_t|\yvect_{1:t})$ (filtering), but as well as when $p(\xvect_{1:T}|\yvect_{1:T})$ (smoothing) is evaluated. While the former is used for on-line \textit{prediction}, i.e.\ when the model is already trained, the latter is part of the E-step of any algorithm used for learning the parameters, and therefore equally essential. In the following, we present our variational approximation to perform inference as well as the complete VEM algorithm for learning. We also discuss the derivation of the GPB2 algorithm~\cite{bar1988tracking} in the context of the proposed formulation.

\section{Variational Inference and Learning}
\label{sec:estimation}
In this section we present a variational approximation of P-LDS and we derive a variational EM algorithm with tractable inference (expectation) and closed-form parameter learning (maximization). We assume that the continuous and discrete variables are independent, a posteriori: consequently, the joint distribution over $\xvect_{1:T}$ and $z_{1:T}$ is approximated with the following factorization:
\begin{equation}
\label{eq:var-factorization}
 p(\xvect_{1:T},z_{1:T}|\yvect_{1:T}) \approx q(\xvect_{1:T}) q(z_{1:T}).
\end{equation}
This follows the same philosophy as the factorial hidden Markov models~\cite{ghahramani1996factorial}. However, here we deal with hybrid states, namely discrete and continuous, therefore the derivation~\cite{ghahramani1996factorial} does not apply and we need to derive a new EM algorithm. Notice that the proposed model is different than the model of \cite{ghahramani2000variational}. Indeed, the latter switches between several continuous states, with their own linear dynamic regimes, while the proposed model approximates a possibly non-linear dynamic regime with a piecewise-linear model, similar to GPB2, i.e. \eqref{eq:predictive-gpb2-K1}.

As with HMM and LDS learning, we assume that the entire sequence of observations is available for training and the challenge consists of inferring the entire chain of state variables and of estimating the model parameters, namely the parameter vectors $\thetavect$, i.e. \eqref{eq:theta} (observation model) and $\phivect$, i.e. \eqref{eq:phi} (dynamic model). As already outlined in Section~\ref{sec:temporal}, the parameters $\thetavect$ don't depend on time and they can be estimated using the algorithm of\cite{DeleforgeForbesHoraud2015}. Therefore, we only need to estimate the dynamic parameters $\phivect$.
Formally, we maximize the expected complete-data log-likelihood:
\begin{equation}
 {\cal Q}(\phivect) = \mathbb{E}_{p(\xvect_{1:T},z_{1:T}|\yvect_{1:T})} \big[ \log p(\xvect_{1:T},z_{1:T},\yvect_{1:T}|\phivect)\big],\label{eq:q-aux-func}
\end{equation}
where the posterior distribution $p(\xvect_{1:T},z_{1:T}|\yvect_{1:T})$ is evaluated with the model parameters at the previous iteration $\phivect^\textrm{old}$, implicit in the previous equation to simplify the reading.

\subsection{Inference: The Expectation Steps}

The two posterior distributions \eqref{eq:var-factorization} write:
\begin{align}
\label{eq:qz} 
 \log q(z_{1:T}) &= \mathbb{E}_{q(\xvect_{1:T})} \big[ \log p(\xvect_{1:T},z_{1:T}|\yvect_{1:T})\big] +\textrm{const}, \\
 \label{eq:qx}
  \log q(\xvect_{1:T}) &= \mathbb{E}_{q(z_{1:T})} \big[ \log p(\xvect_{1:T},z_{1:T}|\yvect_{1:T})\big] +\textrm{const}, 
\end{align}
\addnote[alternate]{1} {These distributions are alternatively evaluated, as explained in detail below.}
\subsubsection{E-Z step}
By developing~(\ref{eq:qz}), ignoring the constant terms and using \eqref{eq:dynamic}-\eqref{eq:switching} we obtain:
\begin{align}
 \nonumber q(z_{1:T}) \propto & \; {\cal N} ( \yvect_1;\Amat_{z_1}\etavect_1+\bvect_{z_1},\Sigmamat_{z_1} ) \\
 \nonumber \times &  \exp\left(-\textstyle\frac{1}{2} \textrm{tr}\left(\Amat_{z_1}^\top \Sigmamat_{z_1}\inverse \Amat_{z_1}\Vmat_1\right) \right) \\
 \nonumber \times &  {\cal N}(\etavect_1;\gammavect_{z_1},\Gammamat_{z_1}) \exp\left(-\textstyle\frac{1}{2} \textrm{tr}\left(\Gammamat_{z_1}\inverse \Vmat_1\right)\right) \\
 \nonumber  \times & \prod_{t\geq 2}\bigg( {\cal N}( \yvect_t;\Amat_{z_t}\etavect_t+\bvect_{z_t},\Sigmamat_{z_t} ) \\
 \nonumber  & \times  \exp\left(-\textstyle\frac{1}{2} \textrm{tr}\left(\Amat_{z_t}^\top \Sigmamat_{z_t}\inverse \Amat_{z_t}\Vmat_t\right) \right) \\
 \nonumber & \times   {\cal N}( \etavect_t;\Cmat_{z_t}\etavect_{t-1},\Qmat_{z_t} ) \\
 \nonumber  & \times  \exp\left(-\textstyle\frac{1}{2} \textrm{tr}\left(\Cmat_{z_t}^\top \Qmat_{z_t}\inverse \Cmat_{z_t}\Vmat_{t-1} + \Qmat_{z_t}\inverse\Vmat_t \right) \right) \\
& \times  \exp\left(\textrm{tr}\left( \Qmat_{z_t}\inverse\Cmat_{z_t}\Wmat_{t}\right)\right)\; \tau_{z_{t-1}z_t}\bigg), \label{eq:qz_dev}
\end{align}
where $\etavect_t = \mathbb{E}_{q(\xvect_{1:T})}[\xvect_t]$ is the posterior mean of $\xvect_t$, $\Vmat_t=\mathbb{E}_{q(\xvect_{1:T})}[\xvect_t\xvect_t^\top]-\etavect_t\etavect_t^\top$ is the posterior covariance of $\xvect_t$ and $\Wmat_t=\mathbb{E}_{q(\xvect_{1:T})}[\xvect_{t-1}\xvect_t^\top]-\etavect_{t-1}\etavect_t^\top$ is the posterior cross-covariance of $\xvect_{t-1}$ and $\xvect_t$.

One may notice that it is possible to group the terms that depend on $z$ in \eqref{eq:qz_dev}, thus yielding:
\begin{equation}
 q(z_{1:T}) \propto \iv{\rho}_{1,z_1} \prod_{t\geq 2} \Big( \iv{\rho}_{t,z_t} \tau_{z_{t-1}z_t} \Big),
\end{equation}
which is equivalent to an HMM with observation probabilities $\iv{\rho}_{t,z_t}$. Therefore, by computing the standard forward-backward algorithm for HMMs, one can easily obtain the forward $\fv{\rho}_{t,z_t}$ and backward $\bv{\rho}_{t,z_t}$ probabilities to eventually obtain the posteriori probability $p(z_t|\yvect_{1:T}) \approx q(z_t)$:
\begin{equation}
 q(z_t) = \rho_{t,z_t} \propto  \frac{ \fv{\rho}_{t,z_t}\,\bv{\rho}_{t,z_t} }{ \sum_z \fv{\rho}_{t,z}\,\bv{\rho}_{t,z}}.
\end{equation}
The estimation of the transition parameters $\tau_{zz'}$ requires the joint posteriori probability distribution of $z_{t-1},z_t$ which can be easily computed as:
\begin{equation}
 q(z_{t-1},z_t) = \jv{\rho}_{t,z_{t-1}z_t} \propto \fv{\rho}_{t-1,z_{t-1}} \tau_{z_{t-1}z_t} \iv{\rho}_{t,z_t} \bv{\rho}_{t,z_t}.
\end{equation}

\subsubsection{E-X step}
By developing~(\ref{eq:qx}), ignoring the constant terms and using \eqref{eq:dynamic}-\eqref{eq:switching} we obtain:
\begin{align}
 \nonumber  & \log q(\xvect_{1:T})\propto \\
 \nonumber &  -\textstyle\frac{1}{2}\Big( \xvect_1^\top {(\iv{\Vmat}_1)}\inverse \xvect_1 -2\xvect_1^\top {(\iv{\Vmat}_1)}\inverse \iv{\etavect}_1 \\
 \nonumber & + \xvect_1^\top (\overline{\Gammamat})\inverse \xvect_1 \!-\! 2\xvect_1^\top (\overline{\Gammamat})\inverse \overline{\gammavect} \\
\nonumber & + \sum_{t\geq 2}\big( \xvect_t^\top (\iv{\Vmat}_t)\inverse \xvect_t 
 - 2 \xvect_t^\top (\iv{\Vmat}_t)\inverse \iv{\etavect}_t  + \xvect_t^\top (\overline{\Qmat}_t)\inverse \xvect_t \\
 &  - 2 \xvect_t^\top \overline{\Rmat}_t \xvect_{t-1} + \xvect_{t-1}^\top (\overline{\Smat}_t)\inverse \xvect_{t-1} \big)\Big), \label{eq:qx-dev}
\end{align}
where the following definitions hold:
\begin{align}
\label{eq:var-mean}
 \iv{\etavect}_t &= \iv{\Vmat}_t\Big( \sum_z \rho_{t,z}\Amat_z^\top\Sigmamat_z\inverse(\yvect_t-\bvect_z) \Big), \\
 (\iv{\Vmat}_t)\inverse &= \sum_z \rho_{t,z} \Amat_z^\top \Sigmamat_z\inverse \Amat_z, \\
 \overline{\gammavect} &= \overline{\Gammamat} \sum_z \rho_{1,z} \Gammamat_z\inverse \gammavect_z, \\
 (\overline{\Gammamat})\inverse &= \sum_z \rho_{1,z} \Gammamat_z\inverse, \\
 (\overline{\Qmat}_t)\inverse &= \sum_z \rho_{t,z}\Qmat_z\inverse, \\
 \overline{\Rmat}_t &= \sum_z \rho_{t,z}\Qmat_z\inverse\Cmat_z, \\
 (\overline{\Smat}_t)\inverse &= \sum_z \rho_{t,z}\Cmat_z^\top\Qmat_z\inverse\Cmat_z.
\end{align}
To be valid, the last equation requires that $\rho_{t,z}\Cmat_z^\top\Qmat_z\inverse\Cmat_z$ is invertible for all values of $z$, this implies that $\Cmat_z$ is a full rank matrix for all values of $z$, which is a very mild assumption. This is very close to a standard LDS (Kalman filter), but different enough in that standard forward-backward recursions cannot be applied. Indeed, in a standard LDS the following relationship holds:
$\overline{\Rmat}_t^\top {\overline{\Qmat}_t}\inverse \overline{\Rmat}_t = {\overline{\Smat}_t}\inverse$, which is not the case in general. This condition is equivalent to impose the same Gaussian dynamic model to all the realizations of $Z_t$ (which clearly corresponds to a Kalman filter). However, from~(\ref{eq:qx-dev}), one may easily see that the joint posterior distribution is a high-dimensional Gaussian, and therefore the marginals will also be Gaussian. Even if the relationship does not correspond to a standard LDS, it is still possible to write the forward-backward equations that efficiently solve in an exact manner the inference of $q(\xvect_t)$.

With the above notations the forward and backward recursions write, respectively:
\begin{align*}
 \fv{\etavect}_t &= \fv{\Vmat}_t \left( (\iv{\Vmat}_t)\inverse\iv{\etavect}_t + \overline{\Rmat}_t\overline{\Smat}_t(\overline{\Smat}_t+\fv{\Vmat}_{t-1})\inverse\fv{\etavect}_{t-1} \right), \\
(\fv{\Vmat}_t)\inverse &= (\iv{\Vmat}_t)\inverse + (\overline{\Qmat}_t)\inverse-\overline{\Rmat}_t( (\overline{\Smat}_t)\inverse + (\fv{\Vmat}_{t-1})\inverse )\inverse\overline{\Rmat}_t^\top,
\end{align*}
and:
\begin{align*}
\bv{\etavect}_t &= \bv{\Vmat}_t \overline{\Rmat}_t \Omegamat_{t+1}\inverse \left( (\iv{\Vmat}_{t+1})\inverse\etavect_{t+1}^\iota + (\bv{\Vmat}_{t+1})\inverse\etavect_{t+1}^\leftarrow  \right), \\
(\bv{\Vmat}_t)\inverse &= \overline{\Smat}_{t+1}\inverse - \overline{\Rmat}_{t+1}^\top \Omegamat_{t+1}\inverse \overline{\Rmat}_{t+1}, \\
\Omegamat_{t+1}\inverse &= (\iv{\Vmat}_{t+1})\inverse + \overline{\Qmat}_{t+1}\inverse + (\bv{\Vmat}_{t+1})\inverse.
\end{align*}
The forward is initialized with $(\fv{\Vmat}_1)\inverse = {\overline{\Gammamat}}\inverse + (\iv{\Vmat}_1)\inverse$ and $\fv{\etavect}_1 = \fv{\Vmat}_1\left( {\overline{\Gammamat}}\inverse\overline{\gammavect} + (\iv{\Vmat}_1)\inverse\iv{\etavect}_1\right)$. The backward recursion starts at $t=T-1$ with $(\bv{\Vmat}_{T})\inverse = \zeromat$ (and therefore the value of $\bv{\etavect}_T$ has no effect).
Together, they allow us to write the posterior probability of $\xvect_t$:
\begin{align}
  q(\xvect_t|\yvect_{1:T}) &= {\cal N}(\xvect_t;\etavect_t,\Vmat_t),  \label{eq:posterior-vem}\\
 \nonumber \etavect_t &= \Vmat_t \left( (\bv{\Vmat}_t)\inverse \bv{\etavect}_t + (\fv{\Vmat}_t)\inverse \fv{\etavect}_t \right),\\
 \nonumber (\Vmat_t)\inverse &= (\bv{\Vmat}_t)\inverse + (\fv{\Vmat}_t)\inverse. 
\end{align}

In order to estimate the parameters of the dynamics, $\Cmat_z$ and $\Qmat_z$, one needs the joint posterior probability of $\xvect_{t}$ and $\xvect_{t-1}$:
\begin{align}
 q(\xvect_t,\xvect_{t-1}) &= {\cal N} \left( \xvect_t, \xvect_{t-1} ; \jv{\etavect}_t, \jv{\Vmat}_t \right),\\
\nonumber  \jv{\etavect}_t &= \jv{\Vmat}_t \left(\begin{array}{c} (\iv{\Vmat}_t)\inverse\iv{\etavect}_t + (\bv{\Vmat}_t)\inverse\bv{\etavect}_t \\ (\fv{\Vmat}_{t-1})\inverse\fv{\etavect}_{t-1} \end{array}\right), \\
\nonumber  (\jv{\Vmat}_t)\inverse &=\left( \begin{array}{rc} \Omegamat_t\inverse & -\overline{\Rmat}_t \\ -\overline{\Rmat}_t^\top & \overline{\Smat}_{t}\inverse + (\fv{\Vmat}_{t-1})\inverse \end{array} \right),
\end{align}
from which we compute the matrix $\Wmat_t$, required in~(\ref{eq:qz_dev}), by taking the upper-right block of $\jv{\Vmat}_t$.

\subsection{Learning: The Maximization Step}
The estimation of the dynamic parameters is carried out by taking the derivative of the auxiliary function (\ref{eq:q-aux-func}) with respect $\phivect$. Given the formulas derived in the previous section, the terms of the auxiliary function that depend on $\phivect$ are:
\begin{align}
 {\cal Q}(\phivect) &= \sum_{t\geq 2} \mathbb{E}_{q(\xvect_t,\xvect_{t-1})q(z_t)} \big[ \log p(\xvect_t|\xvect_{t-1},z_t)\big]\nonumber\\
 &+ \sum_{t\geq 2}\mathbb{E}_{q(z_t,z_{t-1})} \big[ \log p(z_t|z_{t-1}) \big].
\end{align}
Taking the expectation with respect to all probabilities, including the discrete state variables $Z_{t}$, and using the dynamic models, we obtain:
\begin{align}
 \nonumber & {\cal Q}  (\phivect) = \frac{1}{2} \sum_{t\geq 2}\sum_z \rho_{tz} \int_{\xvect_t,\xvect_{t-1}} \!\!\!\!\!\!\!\!\!\!\!\! {\cal N} ( (\xvect_t;\xvect_{t-1}); \jv{\etavect}_t,\jv{\Vmat}_t) \\
 \nonumber& \times \Big( \log |\Qmat_z\inverse| - (\xvect_t-\Cmat_z\xvect_{t-1})^\top \Qmat_z\inverse (\xvect_t-\Cmat_z\xvect_{t-1})\Big) \textrm{d}\xvect_t\xvect_{t-1} \\
 & + \sum_{t\geq 2}\sum_{z,z'} \rho_{t,zz'}\log \tau_{zz'}.
\end{align}
Taking the expectation with respect to the continuous variables $\xvect_t$, we obtain:
\begin{align}
 \nonumber{\cal Q}(\phivect) &= \frac{1}{2} \sum_{t\geq 2}\sum_z \rho_{tz} \Big( \log|\Qmat_z\inverse| \\
 \nonumber &- (\etavect_t-\Cmat_z\etavect_{t-1})^\top \Qmat_z\inverse (\etavect_t-\Cmat_z\etavect_{t-1})\\
 \nonumber &-\textrm{tr}(\Qmat_z\inverse(\Cmat_z\Vmat_{t-1}\Cmat_z^\top + \Vmat_t-2\Cmat_z\Wmat_t))\Big) \\
 & + \sum_{t\geq 2} \sum_{z,z} \rho_{t,zz'} \log \tau_{zz'}.
\end{align}
The optimal values of the transition parameters correspond to a standard HMM model and are given by:
\begin{equation}
 \tau_{zz'} = \frac{1}{T-1}\sum_{t\geq 2} \rho_{t,zz'}.\label{eq:update-tau}
\end{equation}
The optimal value of $\Qmat_z$ is obtained by taking the derivative of ${\cal Q}$ with respect to $\Qmat_z\inverse$ and setting this derivative equal to zero:
\begin{align}
 \nonumber \Qmat_z & = \frac{1}{\sum_{t\geq 2}\rho_{tz}}  \sum_{t\geq 2} \rho_{tz} \Big( (\etavect_t-\Cmat_z\etavect_{t-1})(\etavect_t-\Cmat_z\etavect_{t-1})^\top \\
 & + \Cmat_z\Vmat_{t-1}\Cmat_z^\top + \Vmat_t - 2\Cmat_z\Wmat_t\Big).\label{eq:update-qmat}
\end{align}
Taking the derivative of ${\cal Q}$ with respect to $\Cmat_z$ is more involved since one needs to take the derivative of the matrix-trace operator. After setting the derivatives equal to zero we obtain:
\begin{align}
 \label{eq:update-cmat} \Cmat_z = &\sum_{t\geq 2} \rho_{tz} (\Wmat_t^\top + \etavect_t\etavect_{t-1}^\top) 
 \Big( \sum_{t\geq 2}\rho_{tz}(\Vmat_{t-1}+\etavect_{t-1}\etavect_{t-1}^\top) \Big)\inverse
\end{align}

\section{GPB2 Inference and Learning}
\label{section:gpb2}
GPB2 \cite{bar1988tracking} is a commonly used algorithm to deal with the intractability of  switching LDSs. 
For the sake of completeness, we describe the GPB2 algorithm for the particular case of P-LDS.
As above, only the dynamic parameters $\phivect$, i.e. \eqref{eq:phi}, need to be estimated.
The central idea of GPB2 is to recursively collapse a $K^2$-component mixture into a $K$-component mixture, as explained below. This implies that at each time index $t$ the conditional posterior $p(\xvect_t | \yvect_{1:t} )$ is successively approximated. To do that, the marginalization chain in (\ref{eq:predictive-marginal}) is truncated, yielding:
\begin{align}
\label{eq:predictive-gpb2}
p(\xvect_t | \yvect_{1:t} )= 
\sum_{z_{t-1}=1}^K \sum_{z_{t}=1}^K
\int_{\xvect_{t-1}} p(\xvect_{t-1}, \xvect_t, z_{t-1},z_t | \yvect_{1:t}) d \xvect_{t-1},
\end{align}
with:
\begin{align}
\label{eq:joint-posterior-t-1}
& p(\xvect_{t-1}, \xvect_t , z_{t-1}, z_t | \yvect_{1:t})  \propto   \\
& p(\yvect_t | \xvect_t, z_t) p(\xvect_t | \xvect_{t-1}, z_t)   p (z_t | z_{t-1}) p (\xvect_{t-1}, z_{t-1} | \yvect_{1:t-1}).  \nonumber
\end{align}
GPB2 yields the following approximation:
\begin{align}
\label{eq:gpb2-approx}
p (\xvect_{t-1}, z_{t-1} |  & \yvect_{1:t-1}) \\
& \approx \fg{\rho}_{t-1,z_{t-1}} \mathcal{N} ( \xvect_{t-1} | \fg{\etavect}_{t-1,z_{t-1}}, \fg{\Vmat}_{t-1,z_{t-1}}), \nonumber
\end{align}
where $\fg{\rho}_{t,z_{t-1}}$, $\fg{\etavect}_{t,z_{t-1}}$ and $\fg{\Vmat}_{t,z_{t-1}}$ play the same role as $\fv{\rho}_{t,z_{t-1}}$, $\fv{\etavect}_{t,z_{t-1}}$ and $\fv{\Vmat}_{t,z_{t-1}}$ in the previous section, but they take different numerical values because they are computed using the GPB2 approximation instead of the variational one.

By using \eqref{eq:dynamic}, \eqref{eq:observation}, \eqref{eq:switching} and \eqref{eq:joint-posterior-t-1}, the approximation in \eqref{eq:gpb2-approx} boils down to the following relationship:
\begin{align}
\label{eq:predictive-gpb2-K2}
p(\xvect_t | \yvect_{1:t} ) & \approx \sum_{z_{t-1}=1}^K \sum_{z_{t}=1}^K \fg{\rho}_{t,z_{t-1}z_{t}}  \mathcal{N} ( \xvect_{t-1} | \fg{\etavect}_{t,z_{t-1}z_{t}}, \fg{\Vmat}_{t,z_{t-1}z_{t}}),
\end{align}
where the priors $\fg{\rho}_{t,z_{t-1}z_{t}}$, the means $\fg{\etavect}_{t,z_{t-1}z_{t}}$, and the covariances $\fg{\Vmat}_{t,z_{t-1}z_{t}}$ are given by:
\begin{align}
\label{K2-prior}
	\fg{\rho}_{t,z_{t-1}z_{t}} & =  \fg{\rho}_{t,z_{t-1}}\tau_{z_{t-1}z_t} \mathcal{N}\left(\dvect_{t,z_{t-1}z_{t}};0, \Smat_{t,z_{t-1}z_{t}}\right),\\
	\label{K2-mean}
	\fg{\etavect}_{t,z_{t-1}z_{t}} & =  \fg{\Vmat}_{t,z_{t-1}z_{t}} \\
	& \times \Big( \Amat_{z_t}^\top \Sigmamat_{z_{t}}^{-1}\left(\yvect_t - \bvect_{z_t}\right) + \Pmat_{t,z_{t-1}z_{t}}\Cmat_{z_t}\fg{\etavect}_{t,z_{t-1}}\Big) \nonumber \\ 
	\label{K2-covariance}
	\fg{\Vmat}_{t,z_{t-1}z_{t}} & = \Big( \Amat_{z_t}^\top\Sigmamat_{z_t}^{-1}\Amat_{z_t} + \Pmat_{t,z_{t-1}z_{t}}\Big)^{-1},
\end{align}
with:
\begin{align*}
	\dvect_{t,z_{t-1}z_{t}} &= \yvect_t - \Amat_{z_t} (\Cmat_{z_t} \fg{\etavect}_{t,z_{t-1}} ) - \bvect_{z_t},\\
	\Smat_{t,z_{t-1}z_{t}} &= \Sigmamat_{z_t} + \Amat_{z_t} (\Qmat_{z_t} + \Cmat_{z_t} \fg{\Vmat}_{t,z_{t-1}} \Cmat_{z_t}\tp )\Amat_{z_t}\tp,\\
	\Pmat_{t,z_{t-1}z_{t}} &= \left( \Qmat_{z_t} +  \Cmat_{z_t} \fg{\Vmat}_{t,z_{t-1}} \Cmat_{z_t}\tp \right)^{-1}.
\end{align*}
Consequently, the dynamic model expands the $K$-component GMM hypothesized in~(\ref{eq:gpb2-approx}) into a $K^2$-component GMM. GPB2 collapses this $K^2$ components into $K$ components by moment matching, thus obtaining:
\begin{equation}
\label{eq:predictive-gpb2-K1}
p(\xvect_t | \yvect_{1:t} )  \approx \sum_{z_{t}=1}^K \fg{\rho}_{t,z_{t}} \mathcal{N} ( \xvect_{t} | \fg{\etavect}_{t,z_{t}}, \fg{\Vmat}_{t,z_{t}}),
\end{equation}
where $\fg{\rho}_{t,z_{t}}, \fg{\etavect}_{t,z_{t}}, \fg{\Vmat}_{t,z_{t}}$ are given by:
\begin{align}
\fg{\rho}_{t,z_{t}} &= \sum_{z_{t-1}=1}^K  \fg{\rho}_{t,z_{t-1}z_{t}},\\
\fg{\etavect}_{t,z_{t}} &= \sum_{z_{t-1}=1}^K \frac{\fg{\rho}_{t,z_{t-1}z_{t}}}{\fg{\rho}_{t,z_t}} \fg{\etavect}_{t,z_{t-1}z_{t}},\label{gpb2_filt}\\
\fg{\Vmat}_{t,z_{t}} &= \sum_{z_{t-1}=1}^K \frac{\fg{\rho}_{t,z_{t-1}z_{t}}}{\fg{\rho}_{t,z_{t-1}}} \\
 & \times \big( \fg{\Vmat}_{t,z_{t-1}z_{t}}\!\!\!\!+ (\fg{\etavect}_{t,z_{t-1}z_{t}}\!\! \!\!- \fg{\etavect}_{t,z_{t}} )(\fg{\etavect}_{t,z_{t-1}z_{t}}\!\! \!\!- \fg{\etavect}_{t,z_{t}} )\tp\big). \nonumber
\end{align}
Therefore, at each time index $t$, the filtering distribution is approximated with a Gaussian mixture with $K$ components, i.e. \eqref{eq:predictive-gpb2-K1}. The same moment matching technique can be recursively applied to the backward (or smoothing) distribution, thus obtaining a $K$-component Gaussian mixture model for $p(\xvect_t|\yvect_{1:T})$. Finally, it is straightforward to apply the moment matching technique to approximate the joint posterior distribution $p(\xvect_t,\xvect_{t-1}|\yvect_{1:T})$ with a Gaussian mixture with $K$ components, so that the estimation of the transition parameters $\phivect$ is done with the same update formulas as in the variational case, see~(\ref{eq:update-tau}),~(\ref{eq:update-qmat}) and~(\ref{eq:update-cmat}), but using the posterior distribution provided by GPB2.


\section{Experimental Validation}
\label{sec:experiments}
\begin{figure*}[h!t!]
	\centering
	\includegraphics[width=0.95\textwidth]{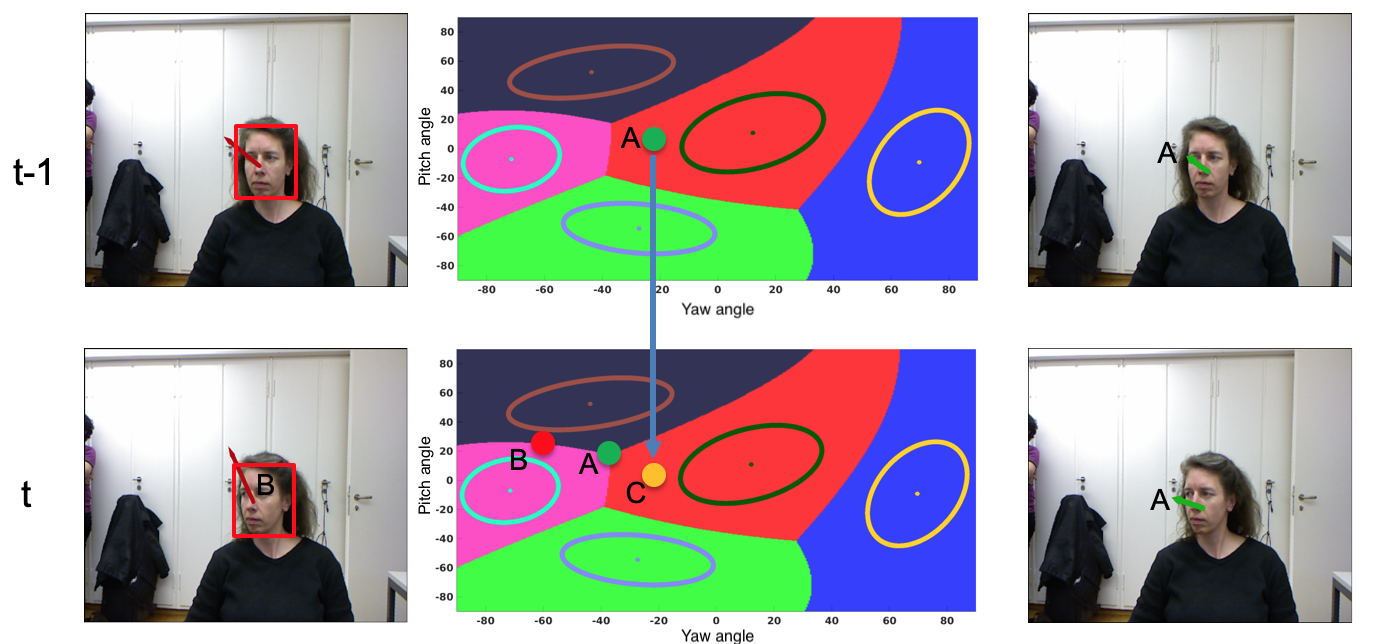}
	\caption{\label{fig:track-principle} 
	The proposed variational P-LDS algorithm applied to the problem of head pose tracking (HPT). The central column shows the Gaussian mixture that models the latent space, i.e. eq. \eqref{eq:model_pXZ}. The parameters of this mixture don't vary over time and they are learnt from a training set of input-output instances of the observed and latent variables. In this example we show the likelihood function associated with the latent variables of head pose, namely the yaw and pitch angles. The observed pose at $t$ (red dot denoted B) is estimated from a high-dimensional feature vector that describes a face (left column). The variational means (green dots denoted A and shown with green arrows onto the right column) are inferred by the E-X step of the algorithm \eqref{eq:var-mean} based on the current dynamic prediction (orange dot denoted C) and the current observation (red dot denoted B).
}
\end{figure*}

In this section we present experimental evaluations of the proposed P-LDS variational EM filtering and smoothing algorithms, namely head pose tracking (HPT) from a video, e.g. Figure~\ref{fig:track-principle}.
The observed data consist of high-dimensional feature vectors, e.g. histogram of oriented gradients (HOG), but any other visual descriptors could be used in practice.
\addnote[separation]{1}{
As already explained in Section~\ref{sec:temporal}, the static parameters can be estimated offline, from a training set of high-dimensional feature vectors (inputs) and the associated ground-truth head-pose angles (outputs). 
In order to choose the number $K$ of linear-Gaussian components, which is equivalent to the number of states of the switch variable, we use the result of \cite{drouard2017robust} which shows that the 
Bayesian information criterion (BIC) model \cite{DeleforgeForbesHoraud2015} may be replaced with an empiric score based on the mean absolute error (MAE) between the predicted head pose and the ground truth. Based on this, we set $K=25$ in all our experiments. It is worthwhile to notice that the off-line training procedure is shared by the variational filter, the variational smoother and GPB2.}

\subsection{Experimental Setup}
\label{sec:hpt}

\def\arraystretch{1.5}
\begin{table}[t!]
	\centering
	\caption{\label{table:datasets} Summary of principal features of the datasets used for empirical evaluation.}	
	\resizebox{1\columnwidth}{!}{	
	\begin{tabular}{p{0.22\columnwidth} p{0.24\columnwidth} p{0.24\columnwidth} p{0.224\columnwidth}}
		\hline
Dataset & Biwi-Kinect \cite{fanelli2013random} & EYEDIAP \cite{mora2014eyediap} & Vernissage \cite{jayagopi2012vernissage} \\
\hline
\#Recordings & 24  & 94  & 10 \\
\#Participants & 20 &16 & 20\\
Sensor type & RGB-D & RGB-D & RGB \\
Pitch range& $[-60^\circ, +60^\circ]$ & $[-40^\circ, +40^\circ]$ & $[-90^\circ, +90^\circ]$ \\
Yaw range & $[-75^\circ, +75^\circ]$ & $[-50^\circ, +50^\circ]$ & $[-90^\circ, +90^\circ]$ \\
Roll range & $[-20^\circ, +20^\circ]$ & N/A & N/A \\
Annotation method & \multicolumn{2}{p{0.48\columnwidth}}{Automatic fitting with a deformable 3D shape model} & Optical motion capture device \\
		\hline
	\end{tabular}
	}
	\end{table}

We empirically evaluate the performance of the proposed methods with three publicly available datasets: Biwi-Kinect \cite{fanelli2013random}, EYEDIAP \cite{mora2014eyediap} and Vernissage \cite{jayagopi2012vernissage} (see~Table~\ref{table:datasets}):
\begin{itemize}
\item The Biwi-Kinect dataset contains 24 videos of 20 people (16 men and 4 women) recorded with a Kinect camera. During the recordings, people were asked to move their heads freely in front of the camera. 3D head pose (pitch, yaw, and roll angles) annotations were obtained automatically and accurately for each video frame using the face-shift software. The angle values range from $-60^{\circ}$ to $60^{\circ}$ for pitch, $-75^{\circ}$ to $75^{\circ}$ for yaw and $-20^{\circ}$ to $20^{\circ}$ for roll. The dataset provides RGB and depth images as well as the calibration matrices. The 3D nose positions are provided as well.
\item The EYEDIAP dataset is intended for both eye-gaze and head-pose estimation. It contains 94 videos of 16 people recorded using different configurations, e.g. static and rotating heads. The dataset provides RGB videos, with both HD and VGA resolution, and depth videos with the associated calibration matrices. Annotations of both head-pose  and eye-gaze are provided for each video frame.  The angle values range from $-40^{\circ}$ to $40^{\circ}$ for pitch and $-50^{\circ}$ to $50^{\circ}$ for yaw. 
	\item The Vernissage dataset contains 10 recordings of 20 people interacting with each and with a robot. Each recording comprises two people. The dataset was recorded with a camera embedded into the head of a robot head. A network of infrared cameras combined with optical markers placed on the participants' heads provide accurate ground-truth head positions and orientations. 
	\end{itemize}

\subsection{Implementation details}

Facial features are computed as follows. A face detector provides bounding boxes for each frame and for each video. Then a high-dimensional feature vector is extracted from each bounding box using  histogram of oriented gradients (HOG) descriptors, obtained along the implementation described in \cite{drouard2017robust} which yields vector-valued observations of dimension $D=1888$. 
For all datasets and for each face identity, we split the corresponding videos into two disjoint sets: a training set and a test set. Since the datasets are annotated, i.e. there are ground-truth head-pose parameter values associated with each frame. As already mentioned, we use the method of \cite{drouard2017robust} to estimate $\thetavect$. 

The dynamic parameters are initialized as follows. First we set $\Cmat_k=\Imat$ since we noticed that the simultaneous estimation of $\Cmat_k$ and $\Qmat_k$ is subject to instabilities in the estimation of the dynamic parameters. The covariance matrices $\left\lbrace \Cmat_j, \Qmat_j\right\rbrace_{j=1}^K$ are initialized with the identity matrix. The entries of the transition matrix 
$\left\lbrace \tau_{ij} \right\rbrace_{i,j = 1}^K$ are initialized in the following way. We compute the pairwise Bhattacharrya distances \cite{bhattachayya1943measure} between the $K$ Gaussian components defined by \eqref{eq:model_pXZ}. The variational EM algorithm alternates between inference of  $q(Z_t)$ (E-Z step) and of $q(\xvect_t)$ (E-X step). The variational parameters $\etavect_t$ and $\Vmat_t$ are initialized with their previously estimated values, namely: 
$\etavect_t = \etavect_{t-1}$ and $\Vmat_t = \Vmat_{t-1}$.

We implemented both the variational filter and the variational smoother described in detail in the previous section. The main difference between the filter and the smoother is the amount of information that is used at inference time. Indeed, while the filter uses only causal information (i.e.\ past observations) the smoother uses also non-causal information (i.e.\ past and future observations). This is why, a priori, one expects the smoother to have better performance, at the price of being a completely off-line algorithm that cannot be used for real-time applications.

\subsection{Results and Discussion}

The proposed methods were compared with the following state-of-the-art HPT methods:
\begin{itemize}
\item \textit{Flandmarks} \cite{uricar2012detector,cech20143d} combines \addnote[fland]{1}{2D face landmark detection with head pose estimation and with tracking. Head pose is estimated in the following way: the 3D landmarks of a mean face are projected onto the image plane and the error between the projected landmarks and the observed landmarks is minimized over the pose parameters. At each time step this non-linear minimizer is initialized with the pose parameters computed at the previous time step. The publicly available implementation of this method only computes the pitch and yaw angles.}
\item \textit{OpenFace} \cite{baltruvsaitis2016openface} is an open source software package for facial behavior analysis, i.e. facial landmark detection and tracking, head pose and eye gaze estimation. It extracts 68 2D facial landmarks using conditional local neural fields and tracks them over time with a three-layer CNN trained to predict landmark detection errors \cite{baltrusaitis2013constrained}. Once 2D landmarks are detected and tracked, they are used in conjunction with a 3D facial model to compute head pose parameters.
\item \textit{ICP tracking} \cite{mora2012gaze} uses both depth and color (RGB-D images) from which 16 3D facial landmarks are manually extracted. The landmarks are tracked based on 
estimating the rigid motion between consecutive frames. 
\end{itemize}

\addnote[2dlandmarks]{1}{These HPT methods make use of 2D facial landmarks, whose detection is known to be sensitive to large head poses that induce partial occlusions of the face. This stays in contrast with the proposed method that directly exploits high-dimensional descriptors of faces}. For completeness, we also compared our algorithms with an implementation of the Kalman filter (LDS), namely \eqref{eq:obs}-\eqref{eq:prior_Z} with $K=1$, with
the HPE method of \cite{drouard2017robust} and with GPB2, i.e. Section~\ref{section:gpb2} and~\cite{drouard2017switching}. To quantitatively evaluate HPT we compute average and standard deviation of the absolute error between the estimated pose parameters and the ground-truth parameters provided with each annotated dataset.

\def\arraystretch{1.5}
\begin{table}[t!]
	\centering
	\caption{\label{table:resultsBIWI} Average (Avg.) and standard deviation (Std.) of the absolute error (in degrees) for the pitch, yaw and roll angles (when applicable) on the Biwi-Kinect dataset. The landmark-based method of \cite{uricar2012detector,cech20143d} only estimates the pitch and the roll angles. The best results are in bold and the second best are in slanted bold. The results that did not pass the Wilcoxon statistical test are marked with an asterisk. The same face detector was used by all methods.}	
	\resizebox{1\columnwidth}{!}{	
	\begin{tabular}{l c c c c c c}
		\hline
		 & \multicolumn{2}{c}{Pitch} & \multicolumn{2}{c}{Yaw} & \multicolumn{2}{c}{Roll}\\
		Method & Avg. & Std. & Avg. & Std. & Avg. & Std.\\
		\hline
		HPE\_GLLiM\_HOG  \cite{drouard2017robust} & $10.54$ & $13.38$ & $11.15$ & $17.93$ & $5.23$ & $5.99$\\	
		HPT\_Flandmarks \cite{uricar2012detector,cech20143d} & $13.12$ & $10.79$ & $21.1$ & $14.16$ & $-$ & $-$\\
		HPT\_OpenFace \cite{baltruvsaitis2016openface} & $9.23$ & $15.69$ & $29.43^\ast$ & $25.74$ & $10.72^\ast$& $11.33$\\
		HPT\_GLLiM\_Kalman & $10.35$ & $13.19$ & \textsl{\textbf{10.97}} & $17.75$ & $5.12$ & $5.93$\\
		HPT\_GPB2\_HOG & $\mathbf{9.03}$ & $10.89$ & $\mathbf{8.77}$ & $\textbf{13.42}$ & $\textbf{4.75}$ & $5.11$\\
		HPT\_VarFilter\_HOG & $9.39$ & \textsl{\textbf{8.95}} & $11.81$ & $14.06$ & $4.96$ & \textsl{\textbf{5.01}}\\
		HPT\_VarSmoother\_HOG & \textsl{\textbf{9.08}} & $\mathbf{8.64}$ & $11.06$ & \textsl{\textbf{13.47}} & \textsl{\textbf{4.87}} & $\textbf{4.95}$\\
		\hline
	\end{tabular}
	}
\end{table}

\begin{figure}[t!]
	\centering
	\begin{tabular}{c}
	\includegraphics[trim=2.cm 0.cm 2cm 0.1cm,clip,width=0.99\columnwidth]{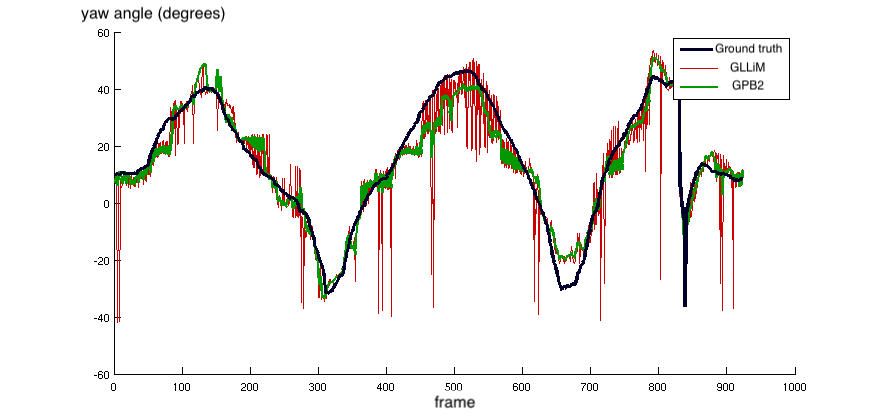}\\
	\includegraphics[trim=2.cm 0.cm 2cm 0.1cm,clip,width=0.99\columnwidth]{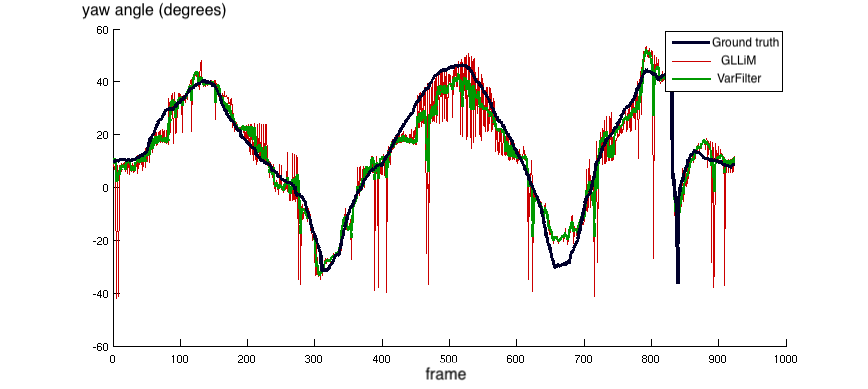}
	\end{tabular}
	\caption{\label{fig:track-biwi} An example from the Biwi-Kinect dataset of a person that rotates his head from left to right and then from right to left (yaw angle). The ground truth trajectory is shown with a blue curve, the result of head-pose estimation (HPE\_GLLiM) is shown in red. The results obtained with the HPT\_GPB2 algorithm  (top) and with the proposed HPT\_VarFilter algorithm (bottom), respectively. Notice that the large errors that are produced by HPE\_GLLiM are eliminated by both trackers.}
\end{figure}

The results obtained with the Biwi-Kinect, EYEDIAP and Vernissage datasets are summarized in tables~\ref{table:resultsBIWI}, \ref{table:resultsEYEDIAP} and \ref{table:resultsVernissage}. 
\addnote[exps-discussion]{1}{The HPT method of \cite{uricar2012detector,cech20143d} does not estimate the roll angle. Moreover, this method relies on 2D landmark detection. Therefore, it fails whenever all the landmarks are not properly detected. This is the reason for which this method did not provide publishable results, when applied to the EYEDIAP and Vernissage datasets. The method of \cite{mora2012gaze} takes as input RGB-D images and manually annotated facial landmarks, hence it could only be applied to the EYEDIAP dataset. In general, the proposed descriptor-based trackers perform better than the landmark-based trackers. A lower standard deviation corresponds to a higher precision and a better method repeatability.}

\addnote[statistical-tests]{1}{In order to assess  the statistical differences in performance between the various methods, we used the Wilcoxon signed-rank test~\cite{wilcoxon1992individual}, inspired from~\cite{lathuiliere2020comprehensive}, which is a non-parametric statistical hypothesis test that is commonly used to compare two related samples in order to assess the null hypothesis that the median difference between pairs of observations is zero. It can be used as an alternative to the paired Student's t-test (t-test for matched pairs) when the population cannot be assumed to be normally distributed. For most of the comparisons, there were no statistical differences. The only notable exception is OpenFace~\cite{baltruvsaitis2016openface}, which performs considerably worse than the other methods on two of the three datasets: the yaw and roll estimation for the Biwi-Kinect dataset, and the pitch and yaw estimation for the EYEDIAP dataset. These estimation are marked with an asterisk in Table~\ref{table:resultsBIWI} and Table~\ref{table:resultsEYEDIAP}, respectively.}

\addnote[figure3]{1}{
As an example, Figure~\ref{fig:track-biwi} shows a ground-truth yaw (left-right rotation) trajectory as well as trajectories obtained with HPE\_GLLiM \cite{drouard2017robust}, with HPT\_GPB2 (top) and with the proposed variational filter (bottom). Overall, the performance of the proposed variational EM algorithms is comparable with the performance of GPB2. Notice that the large errors produced by HPE\_GLLiM are filtered by both trackers}. 

\addnote[CPU]{1}{We measured the CPU time needed to compute one time step. This comprises the following processes: (i)~extraction of a HOG descriptor from a face, (ii) head pose estimation, and (iii)~tracking, where the tracker can be either the GPB2 or the variational filter. Using an Intel-Xeon and Matlab, it takes 1.0 second to extract a HOG descriptor and to estimated head pose, 8.55 seconds to run GPB2 and 2.45 seconds to run the variational filter, respectively.}


\def\arraystretch{1.5}
\begin{table}[t!]
	\centering
	\caption{\label{table:resultsEYEDIAP} Average (Avg.) and standard deviation (Std.) of the absolute error (in degrees) for the pitch and yaw angles on the EYEDIAP dataset. The method of \cite{mora2012gaze} uses both color and depth data and it requires manually annotated 2D landmarks. The best results are in bold and the second best are in slanted bold. The results that did not pass the Wilcoxon statistical test are marked with an asterisk. The same face detector was used by all methods.}
	
	\resizebox{\columnwidth}{!}{	
	\begin{tabular}{l c c c c}
		\hline
		 & \multicolumn{2}{c}{Pitch} & \multicolumn{2}{c}{Yaw}\\
		Method & Avg. & Std. & Avg. & Std.\\
		\hline
		HPE\_GLLiM\_HOG  \cite{drouard2017robust}  & $6.29$ & 7.80 & $\textsl{\textbf{7.80}}$ & $10.39$\\
		ICP tracking \cite{mora2012gaze} & $\textbf{4.17}$ & $\textbf{5.59}$ & $\textbf{6.89}$ & $14.42$\\
		OpenFace \cite{baltruvsaitis2016openface} & $15.39^\ast$ & $12.85$ & $22.21^\ast$ & $16.32$\\
		HPT\_GLLiM\_Kalman & \textsl{\textbf{6.21}} & \textsl{\textbf{7.75}} & $10.62$ & $\textbf{10.31}$\\		
		HPT\_GPB2\_HOG  & $6.68$ & $8.75$ & $8.44$ & \textsl{\textbf{10.91}}\\
		HPT\_VarFilter\_HOG  & $6.96$ & $8.04$ & $11.38$ & $11.44$\\
		HPT\_VarSmoother\_HOG  & $6.78$ & $7.88$ & $10.66$& $10.99$ \\
		\hline
	\end{tabular}
	}
\end{table}

\def\arraystretch{1.5}
\begin{table}[t!]
	\centering
	\caption{Average (Avg.) and standard deviation (Std.) of the absolute error (in degrees) for the pitch and yaw angles on the Vernissage dataset. The best results are in bold and the second best are in slanted bold. The same face detector was used by all methods.}
	\resizebox{\columnwidth}{!}{	
	\begin{tabular}{l c c c c}
		\hline
		 & \multicolumn{2}{c}{Pitch} & \multicolumn{2}{c}{Yaw}\\
		Method & Avg. & Std. & Avg. & Std.\\
		\hline
		HPE\_GLLiM\_HOG  \cite{drouard2017robust}  & $23.95$ & $23.18$ & \textsl{\textbf{11.03}} & $8.57$\\
		OpenFace \cite{baltruvsaitis2016openface} & $21.30$ & \textsl{\textbf{18.80}} & $13.18$ & $10.67$\\
		HPT\_GLLiM\_Kalman & $23.94$ & $23.18$ & \textsl{\textbf{11.03}} & $8.56$\\		
		HPT\_GPB2\_HOG & $\textbf{20.24}$ & $20.62$ & $\textbf{10.21}$ & $\textbf{7.80}$\\
		HPT\_VarFilter\_HOG  & $21.06$ & $19.96$ & $13.76$ & $8.25$\\
		HPT\_VarSmoother\_HOG  & \textsl{\textbf{20.37}} & $\textbf{18.58}$ & $12.92$ & \textsl{\textbf{7.89}}\\
		\hline
	\end{tabular}
	}
	\label{table:resultsVernissage}
\end{table}

\section{Conclusions}
\label{sec:conclusion}
\addnote[conclusions]{1}{In this paper we addressed the problem of learning and inference of piecewise LDS. The latter belongs to the switching LDS class of models, which is known to be intractable because of the combinatorial explosion, over time, of the modes of the latent space. The standard way of dealing with this problem is to use the GPB2 algorithm which collapses a $K^2$-component mixture into a $K$-component one based on moment matching -- a computationally demanding process. Alternatively, we propose a variational approximation of P-LDS and two associated EM algorithms: a variational filter and a variational smoother. Both the filter and the smoother are based on closed-form expressions, which guarantees efficient computation and fast convergence. The proposed variational filter is of the order of 3.5 times faster than the GPB2 method. Not surprisingly, the most time-consuming part of GPB2 is the evaluation of the parameters of the  $K^2$-component Gaussian mixture, followed by the evaluation of the parameters of the approximating $K$-component mixture. This collapsing process resides at the heart of GPB2 and it cannot be avoided. In contrast, the M-step of the proposed VEM can be skipped, once the parameters are learnt, as done in \cite{ba2016line}, which can further accelerate the algorithm.}


\addnote[more-conclusions]{1}{We applied the proposed algorithms to the problem of head-pose tracking.
We presented a series of experiments using several datasets. We carried out a benchmark that included our algorithms and several state-of-the-art tracking algorithms. We note that the variational-based tracker compares well with GPB2 and with the other trackers. It should be noted, however, that the landmark based methods, e.g. \cite{uricar2012detector,cech20143d}, fail to track in many cases. The best performing method is  \cite{mora2012gaze}. Nevertheless, this method has limitations because it requires manual annotation of facial landmarks and it can only be applied to RGB-D images. In contrast, the proposed method is based on extracting high-dimensional descriptors from RGB images, and appears to be more robust against facial self occlusions than the landmark-based methods. }

\balance
\bibliographystyle{IEEEtran}

\end{document}